\documentclass[letterpaper, 10pt, conference]{ieeeconf}      
\usepackage{multirow}
\usepackage{amsmath}
\usepackage{array}
\usepackage{tabu}
\usepackage{graphicx}
\usepackage{amsmath}
\usepackage{scalerel}
\usepackage{amssymb}

\IEEEoverridecommandlockouts                              
\title{\LARGE \bf
Learning to Run with Potential-Based Reward Shaping and Demonstrations from Video Data 
}

\author{Aleksandra Malysheva, Daniel Kudenko and Aleksei Shpilman
\thanks{A. Malysheva is with JetBrains Research, St Petersburg, Russia 
and National Research University Higher School of Economics, St Petersburg, Russia
        {\tt\small malyshevasasha777@gmail.com}}%
\thanks{D. Kudenko is with the Department of Computer Science, University of York, UK, the National Research University Higher School of Economics, St Petersburg, Russia, and JetBrains Research,
        St Petersburg, Russia
        {\tt\small daniel.kudenko@york.ac.uk}}%
\thanks{A. Shpilman is with JetBrains Research, St Petersburg, Russia 
and National Research University Higher School of Economics, St Petersburg, Russia
        {\tt\small aleksei@shpilman.com}}%
}

\begin{document}

\maketitle
\thispagestyle{empty}
\pagestyle{empty}


\begin{abstract}  
Learning to produce efficient movement behaviour for humanoid robots from scratch is a hard problem, as has been illustrated by the "Learning to run" competition at NIPS 2017. The goal of this competition was to train a two-legged model of a humanoid body to run in a simulated race course with maximum speed. All submissions took a tabula rasa approach to reinforcement learning (RL) and were able to produce relatively fast, but not optimal running behaviour. In this paper, we demonstrate how data from videos of human running (e.g. taken from YouTube) can be used to shape the reward of the humanoid learning agent to speed up the learning and produce a better result. Specifically, we are using the positions of key body parts at regular time intervals to define a potential function for potential-based reward shaping (PBRS). Since PBRS does not change the optimal policy, this approach allows the RL agent to overcome sub-optimalities in the human movements that are shown in the videos. 

We present experiments in which we combine selected techniques from the top ten approaches from the NIPS competition with further optimizations to create an high-performing agent as a baseline. We then demonstrate how video-based reward shaping improves the performance further, resulting in an RL agent that runs twice as fast as the baseline in 12 hours of training. We furthermore show that our approach can overcome sub-optimal running behaviour in videos, with the learned policy significantly outperforming that of the running agent from the video.

\end{abstract}


\maketitle


\section{Introduction}

Replicating human movements and behaviour in humanoid robots is a formidable challenge with many exciting applications, ranging from health care (e.g. artificial limbs \cite{alshamsi2016}) to space exploration \cite{nasa2015}. Since the manual engineering of controllers for such tasks is extremely difficult, machine learning and specifically reinforcement learning has received much attention in this area. A recent NIPS competition \cite{kidzinski2018learningtorun} focused on the creation of a simulated humanoid running robot in a continuous and high-dimensional environment. The top entries to the competition employed state-of-the-art deep reinforcement learning techniques \cite{jaskowski2018rltorunfast,kidzinski2018l2rsolutions}, which resulted in strong, but not optimal, performance. 

In this paper, we demonstrate that videos showing human running behaviour can be used to significantly improve the learning performance. In our approach, we use the coordinates of specific body parts (e.g. the foot) to define a potential function that is fed into potential-based reward shaping \cite{ng1999policy}. 

To create a strong baseline for the evaluation of our approach, we combined selected RL techniques of the top ten competition entries with further optimizations to create a running agent that displays a significantly faster learning rate than the top entry. We then add the reward shaping from video data sampled from various YouTube videos, which resulted in a running agent that reached twice the running speed as our baseline in 12 hours of training.

Since potential-based reward shaping has the nice theoretical property of not changing the optimal policy \cite{ng1999policy}, data taken from sub-optimal running behaviour does not prevent the RL agent from overcoming the sub-optimalities and produce humanoid running that outperforms the data source. We demonstrate this theoretical property empirically by sampling limb positions from a slower-running agent and show how our approach generates a running robot, that after a relatively short time of training, starts to run faster than the original agent.

Overall, the main contribution of our work is to demonstrate how data extracted from videos of human movements can be used to significantly speed up the reinforcement learning of humanoid robots. While our work focuses on the training of humanoid running behaviour, the proposed techniques can easily be applied to any other form of humanoid movements. 

\section{Background}
\subsection{Reinforcement Learning}
		
Reinforcement learning is a paradigm which allows agents to learn by reward and punishment from interactions with the environment \cite{sutton1984temporal}. The numeric feedback received from the environment is used to improve the agent's actions. The majority of work in the area of reinforcement learning applies a Markov Decision Process (MDP) as a mathematical model \cite{puterman2014markov}.

An MDP is a tuple $\big(S, A, T, R)$, where $S$ is the state space, A is the action space, $T(s,a,s') = Pr(s'|s,a)$ is the probability that action a in state s will lead to state $s'$, and $R(s, a, s')$ is the immediate reward $r$ received when action $a$ taken in state $s$ results in a transition to state $s'$. The problem of solving an MDP is to find a policy (i.e., mapping from states to actions) which maximises the accumulated reward. When the environment dynamics (transition probabilities and reward function) are available, this task can be solved using policy iteration \cite{bertsekas1995dynamic}.

When the environment dynamics are not available, as with most real problem domains, policy iteration cannot be used. However, the concept of an iterative approach remains the backbone of the majority of reinforcement learning algorithms. These algorithms apply so called temporal-difference updates to propagate information about values of states and/or state-action pairs, $Q(s, a)$ [20]. These updates are based on the difference of the two temporally different estimates of a particular state or state-action value. The Q-learning algorithm is such a method [21]. After each transition, $(s, a) \rightarrow (s', r)$, in the environment, it updates state-action values by the formula:

\begin{equation}
    \label{eq:qlearn}
 Q(s,a) \leftarrow Q(s,a) + \alpha[r + \gamma\max Q(s',a') - Q(s,a)]   
\end{equation}
where $\alpha$ is the rate of learning and $\gamma$ is the discount factor. It modifies the value of taking action $a$ in state $s$, when after executing this action the environment returned reward $r$, and moved to a new state $s'$.

\subsection{Potential Based reward shaping}

The idea of reward shaping is to provide an additional reward representative of prior knowledge to reduce the number of suboptimal actions made and so reduce the time needed to learn \cite{ng1999policy, randlov1998learning}. This concept can be represented by the following formula for the Q-learning algorithm:
\begin{equation} \label{eq:shaping}
    Q(s,a) \leftarrow Q(s,a)+\alpha[r+F(s,s')+\gamma\max Q(s',a') - Q(s,a)]
\end{equation}
where $F(s,s')$ is the general form of any state-based shaping reward.
Even though reward shaping has been powerful in many experiments it quickly became apparent that, when used improperly, it can change the optimal policy \cite{randlov1998learning}. To deal with such problems, potential-based reward shaping was proposed \cite{ng1999policy} as the difference of some potential function $\Phi$ defined over a source s and a destination state $s':F(s,s')=\gamma \Phi(s') - \Phi(s)$ where $\gamma$ must be the same discount factor as used in the agent's update rule (see Equation \ref{eq:qlearn}).		
Ng et al. \cite{ng1999policy} proved that potential-based reward shaping, defined according to Equation \ref{eq:shaping}, guarantees learning a policy which is equivalent to the one learned without reward shaping in both infinite and finite horizon MDPs.

Wiewiora \cite{wiewiora2003potential} later proved that an agent learning with potential-based reward shaping and no knowledge-based Q-table initialization will behave identically to an agent without reward shaping when the latter agent's value function is initialized with the same potential function.				

These proofs, and all subsequent proofs regarding potential-based reward shaping including those presented in this paper, require actions to be selected by an advantage-based policy \cite{wiewiora2003potential}. Advantage-based policies select actions based on their relative differences in value and not their exact value. Common examples include greedy, $\epsilon$-greedy and Boltzmann softmax. 

\subsection{Deep-RL}

In Deep RL, the Q value function is represented as a multi-layer neural network \cite{Goodfellow-et-al-2016}. Deep RL algorithms have been shown to perform strongly on RL tasks which have been infeasible to tackle before. Over recent years, a number of algorithms and optimizations have been proposed, and we have chosen to apply 
the Deep Deterministic Policy Gradient (DDPG) algorithm for our application domain \cite{lillicrap2015continuous}. DDPG has been shown to be effective in continuous action domains where classic reinforcement learning methods struggled.  

Specifically, in the DDPG algorithm two neural networks are used: 
$\mu(S)$ is a network (the {\it actor}) that returns the action vector whose components are the values of the corresponding control signals. 
$Q^w(s, a)$ is a second neural network (the {\it critic}, that returns the $Q$ value, i.e. the value estimate of the action of $a$ in state $s$.

\begin{equation} \label{eq:policy5}
\begin{aligned}
    \nabla_\theta J(\pi_\theta) & =\int_{S}\rho^\pi(s)\int_{A}\nabla_\theta\pi_\theta(a|s)Q^w(s,a) da ds \\
    &=\mathbb{E}_{s\sim \rho^\pi,a\sim\pi_\theta}[\nabla_\theta \log \pi_\theta (a|s)Q^w(s,a)]
\end{aligned}
\end{equation}
where $\theta$ is the parameter vector of the probabilistic policy and $\rho^\pi(s)$ is the probability of reaching state $s$ with policy $\pi$.

For a more complete description of DDPG, see \cite{lillicrap2015continuous}.

\subsection{Reinforcement Learning from Demonstration}

Human expert demonstrations have been demonstrated to improve the learning speed and accuracy of RL agent on a wide range of tasks. Most work in this area (e.g. \cite{suay2016learning, brys2015reinforcement}) focused on the use of state-action recordings as demonstration. This is infeasible in the case of video data, where only state information is available and the demonstration actions are not explicitly provided and often can not be derived either (as is the case with running). More recently, various methods for state-only demonstrations have been proposed (e.g. \cite{peng2018deepmimic, liu2017imitation}). However, all of these methods target the imitation of demonstrations. Our work is employing potential-based reward shaping which uses the demonstrations to speed up the learning, but is also able to overcome any sub-optimalities in the demonstration rather than purely imitating them. 

\section{Simulation Environment} 
\label{sec:domain}

\begin{figure}
\centering
\includegraphics[width=0.40\textwidth]{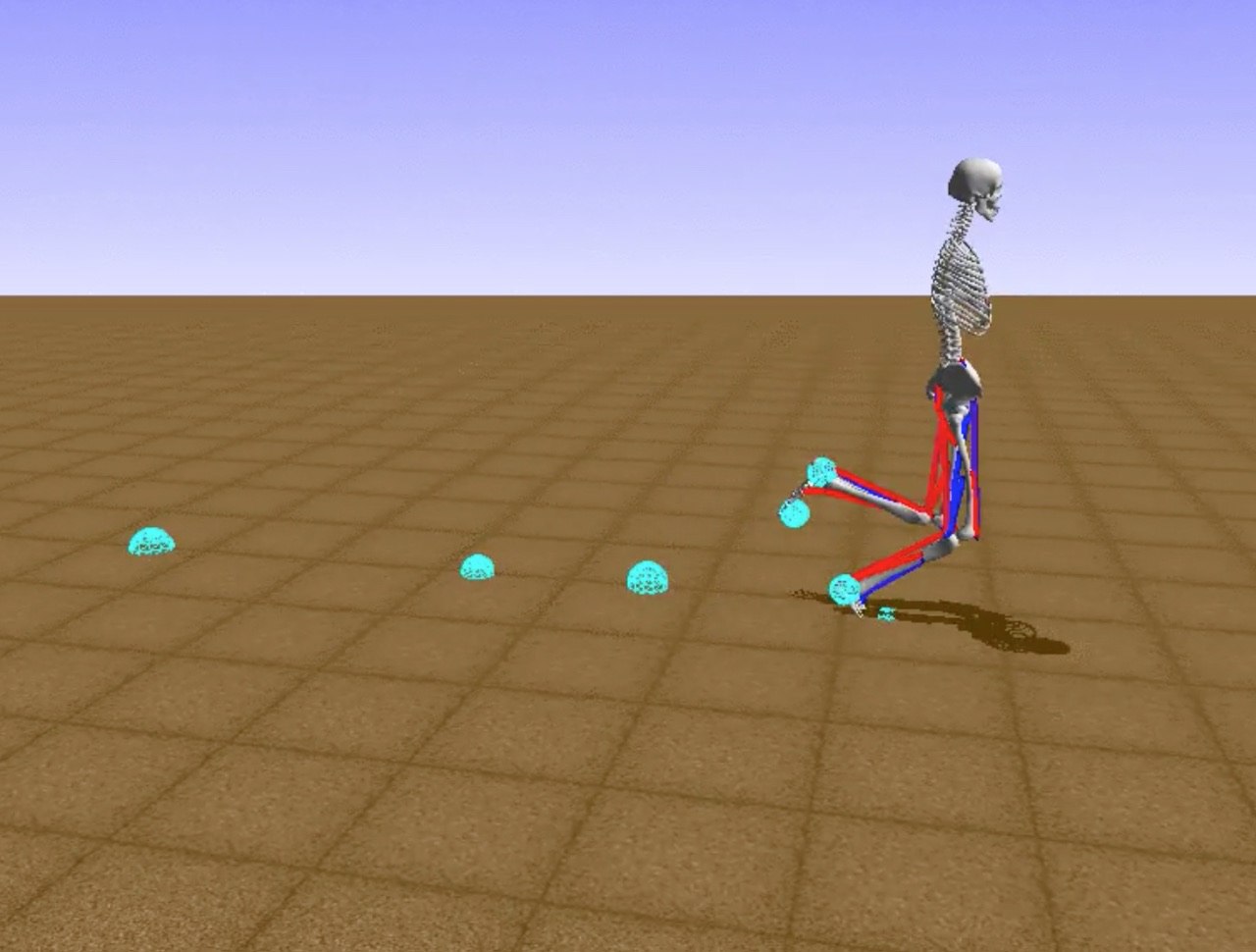}
\caption{A screenshot of the "Learning to Run" simulation environment}
\label{img:environment}
\end{figure}

The simulation environment has been provided by the "Learning to Run" competition and is based on the OpenSim environment employing the Simbody physics engine. The environment simulates a three-dimensional race course with small obstacles, along which a humanoid robot with 6 joints (ankle, knee, and hip on two legs) and corresponding muscles is running (see Figure \ref{img:environment}). The actions of the running humanoid robot are excitation values applied to the muscles implemented in the robot model. The next state of the environment is computed by the physics engine based on the resulting muscle activations, forces, velocities and positions of the joints. 

The OpenSim \cite{kidzinski2018learningtorun} model environment represents the robot state using a vector of 41 features:

\begin{itemize}
    \item position of the pelvis (rotation, x, y)
    \item velocity of the pelvis (rotation, x, y)
    \item rotation of each ankle, knee and hip (6 values)
    \item angular velocity of each ankle, knee and hip (6 values)
    \item position of the center of mass (2 values)
    \item velocity of the center of mass (2 values)
    \item positions of head, pelvis, torso, left and right toes, left and right talus (14 values)
    \item strength of left and right psoas (a muscle at the lower spine)
    \item next obstacle: x distance from the pelvis, y position of the center relative to the the ground, radius.
\end{itemize}

The reward of an agent is provided at each simulation step and is the distance covered in the run minus the muscle strain as computed by the simulation environment. 

\section{Baseline Agent}

When designing the baseline agent, we combined selected techniques from the top 10 competition entries \cite{kidzinski2018l2rsolutions} with further optimizations. In this section, we summarize the most beneficial techniques used, all of which are taken from various contributions published in \cite{kidzinski2018l2rsolutions}.

In all exerimental results presented in the remainder of this paper, the experiments have been repeated 5 times, and the graphs show the standard error from the mean. The RL parameter choice was $\alpha = 0.08$ and $\gamma = 0.9$, which have been determined experimentally.

\subsection{State representation}

The original state representation provided by the competition software contained 41 features, described in Section \ref{sec:domain}. In our state representation we added 71 features, including: 

\begin{itemize}
    \item Two-dimensional coordinates of key body positions relative to the pelvis at the center point (0,0).
    \item Two-dimensional velocity and acceleration vectors for key body points.
\end{itemize} 

\begin{figure}
\includegraphics[width=0.47\textwidth]{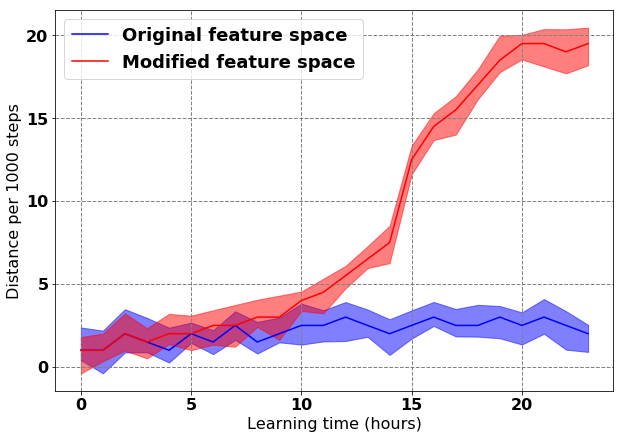}
\caption{This Figure shows significant learning speed increase after adding velocity and acceleration features and centering the coordinates system at the pelvis position}
\label{img:centerandfeatures}
\end{figure}

The new state representation allowed us to significantly speed up the learning process as seen on Figure \ref{img:centerandfeatures}.
 
\subsection{Additional training experience}
\begin{figure}
\includegraphics[width=0.47\textwidth]{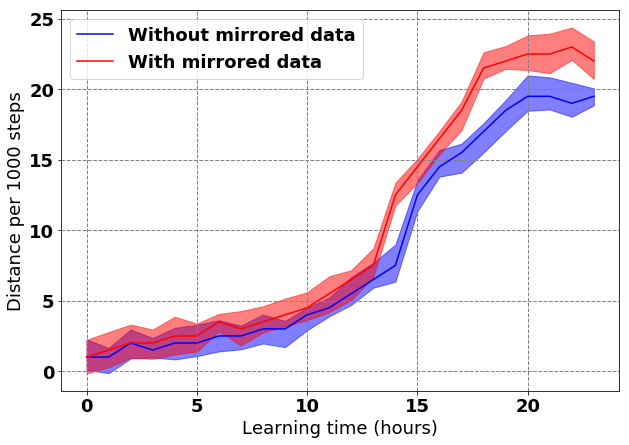}
\caption{Speeding up learning process through adding mirrored data}
\label{img:mirrored}
\end{figure}

After running a simulation episode, we trained the RL agent with additional mirrored data, which represented the agents experience during the episode and reflecting it along the $xy$ plane. This adds valuable training for the value estimator (i.e. the critic), since the task is symmetrical. Figure \ref{img:mirrored} shows the resulting performance improvement. 

\subsection{Repeating the chosen action}
\begin{figure}
\includegraphics[width=0.47\textwidth]{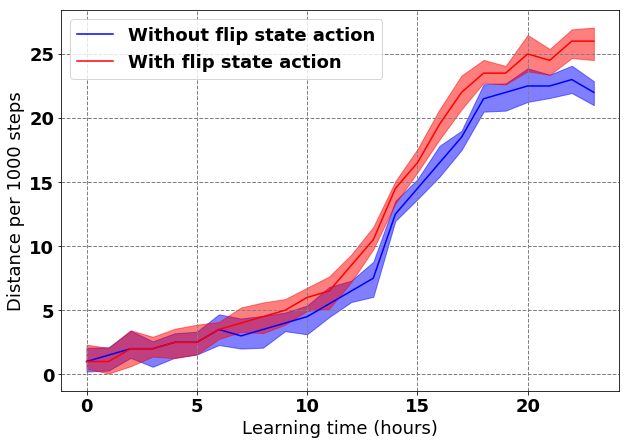}
\caption{Performance gains when repeating each action three times}
\label{img:flipaction}
\end{figure}

Each time the running agent chooses an action, this is repeated three times. Because we employed an actor-critic method, this reduced the number of computations needed to generate the next action during an episode by a factor of three. The resulting performance gain can be seen in Figure \ref{img:flipaction}.

\subsection{Reducing state resolution}

\begin{figure}
\includegraphics[width=0.47\textwidth]{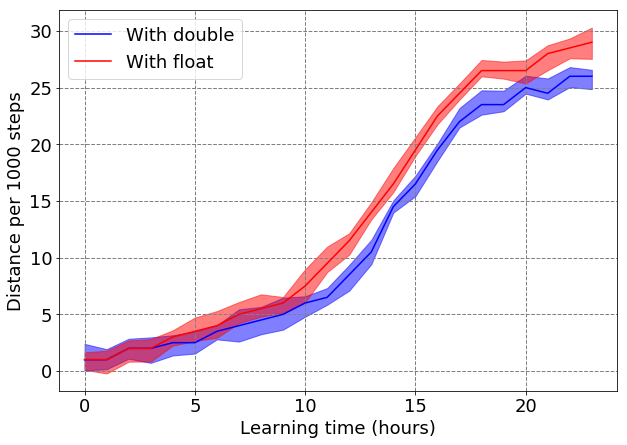}
\caption{Switching from double to float}
\label{img:doublefloat}
\end{figure}

In this optimization step, all the state representation data was changed from {\it double} to {\it float}. This resulted in a speed-up of the computations and a somewhat smaller state space, while also reducing the precision of the state representation. Figure \ref{img:doublefloat} shows the resulting performance increase. 

\subsection{Neural network topology}
\begin{figure}
\includegraphics[width=0.47\textwidth]{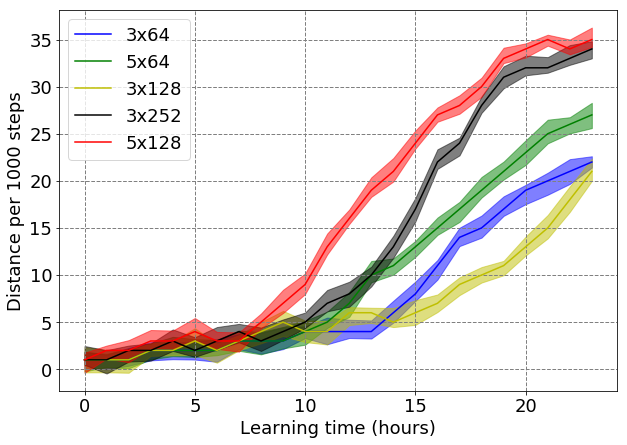}
\caption{Various neural network topologies: LxN denotes L layers with N neurons per layer}
\label{img:architectures}
\end{figure}

After applying all of the techniques above, we compared 5 different network architectures by arbitrarily varying the number of layers and neurons per layer. The results are presented in Figure \ref{img:architectures}. For our baseline agent we chose the best performing layer, using 5 layers with 128 neurons each.

\section{Reward Shaping from Video Data}

After designing our baseline, we added potential-based reward shaping from video data taken from arbitrary YouTube videos depicting running of humans and human-like characters. In this section we describe how the potential function was generated. 

\subsection{Potential function}
The overall potential function is defined as the sum of potential functions for every body part: pelvis, two knees and two feet. Following the potential-based reward shaping approach, an additional reward is given to an agent on each simulation step corresponding to the change in potentials of the source and target state.

We considered the following three different potential functions for each body part (knee and foot) in our research, all of them based on the inverse of the distance between the respective body part coordinate in the video-generated data and the humanoid robot. The three potential functions represent three different inverse distance functions:

\begin{itemize}
    \item PF1: $\frac{1}{dx + dy}$
    \item PF2: $\frac{1}{\sqrt{dx^2 + dy^2}} $
    \item PF3: $\frac{1}{dx^2 + dy^2}$
\end{itemize}

where $dx$ ($dy$) is the absolute difference between the x (y) coordinate of the respective body part taken from the video data and the x (y) coordinate of the body part of the humanoid robot. 

\subsection{Data collection}

\begin{figure}
\centering
\includegraphics[width=0.3\textwidth]{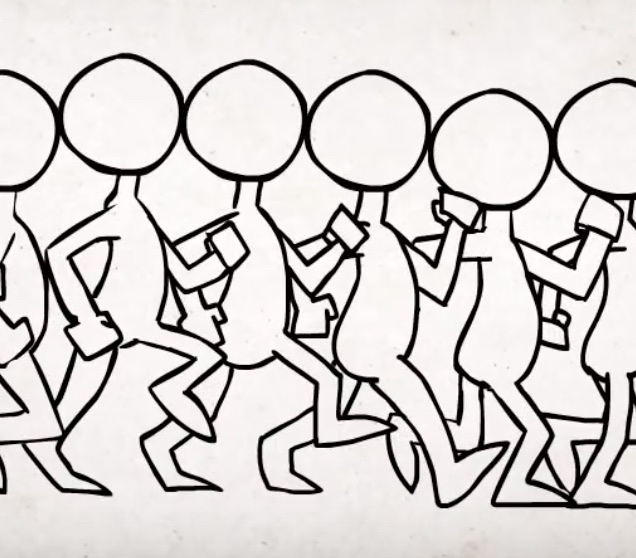}
\caption{Screenshot from a video depicting a cartoon character running (taken from http://y2u.be/2y6aVz0Acx0)}
\label{img:cartoon}
\end{figure}

\begin{figure}
\centering
\includegraphics[width=0.3\textwidth]{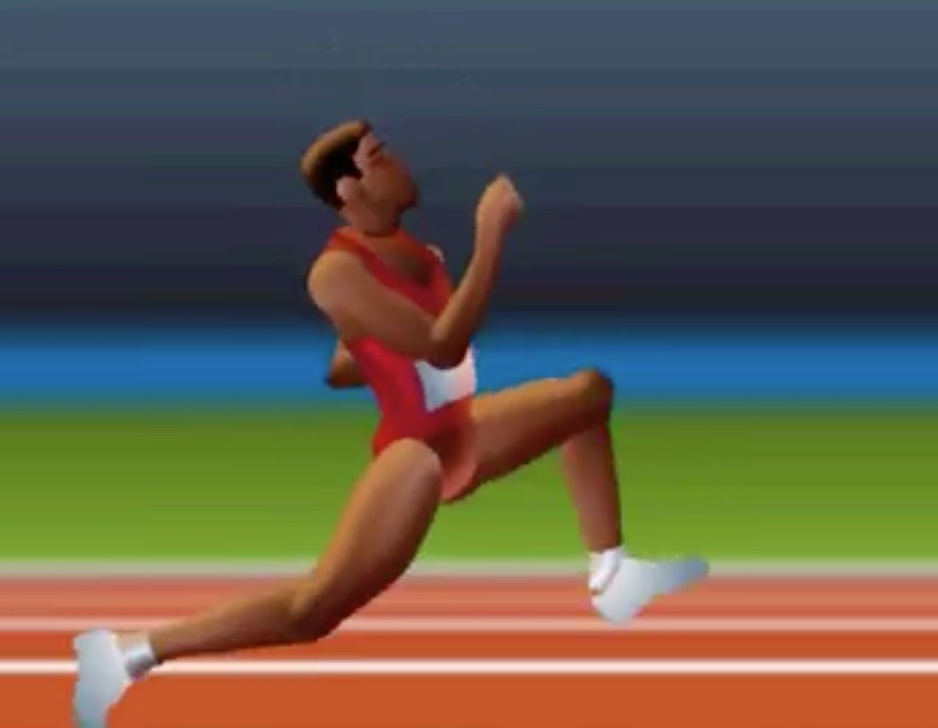}
\caption{Screenshot from a video depicting a computer game character running (taken from http://y2u.be/YbYOsE7JyXs)}
\label{img:videogame}
\end{figure}

\begin{figure}
\centering
\includegraphics[width=0.3\textwidth]{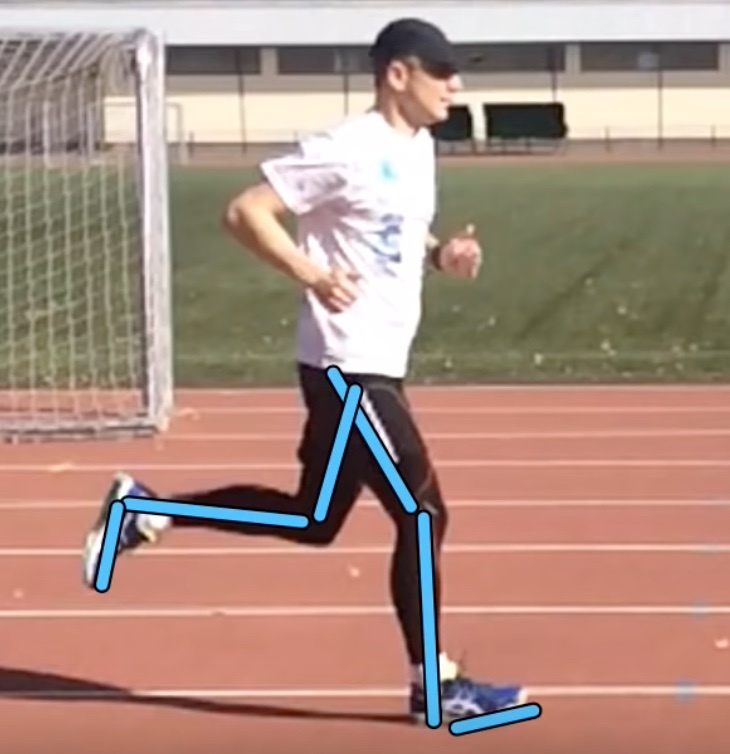}
\caption{Screenshot from a video depicting a human running (taken from http://y2u.be/5mVgThl-yMU)}
\label{img:runninghuman}
\end{figure}

For our potential function we have used the following three sources of video data:

\begin{itemize}
    \item A video of a cartoon character running (see Figure \ref{img:cartoon} for a screenshot)
    \item A video of a running character in a computer game (see Figure \ref{img:videogame} for a screenshot)
    \item A video of a running human (see Figure \ref{img:runninghuman} for a screenshot)
\end{itemize}

Each of the sources was used to define a potential function. The performance of the resulting potential functions in RL are compared in Figure \ref{img:var_potential}. Note that the learning curves were not trained until final convergence, which would require much more time and based on the theoretical properties of potential-based reward shaping would ultimately reach the same performance. In each source we recorded the positions of the two knees and the two feet relative to the pelvis as a two-dimensional coordinate. The recording frequency was four positions per half step. The resulting coordinates were normalized according to the OpenSim simulation. While in our work the extraction of the coordinates was done manually, algorithms to accurately extract body part positions in images with a clear view of the body do exist (e.g. \cite{toshev2014deeppose,guler2018densepose}), and we intend to use these in future work. 

\subsection{Selecting the data source and the potential function}

\begin{figure}
\includegraphics[width=0.47\textwidth]{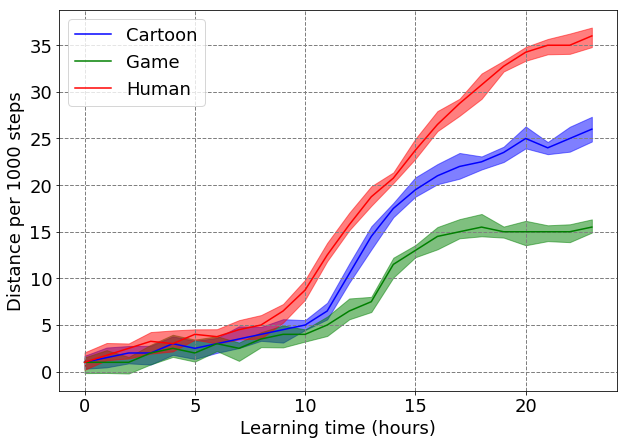}
\caption{Comparing the three videos as a data source for the potential function based on PF2}
\label{img:differentvideo}
\end{figure}

We first compared the performance of the potential functions based on three videos and the inverse distance measure PF2. The results for this experiment is shown in Figure \ref{img:differentvideo}, and demonstrates that the human video is the best data source for the reward shaping. 

\begin{figure}
\includegraphics[width=0.47\textwidth]{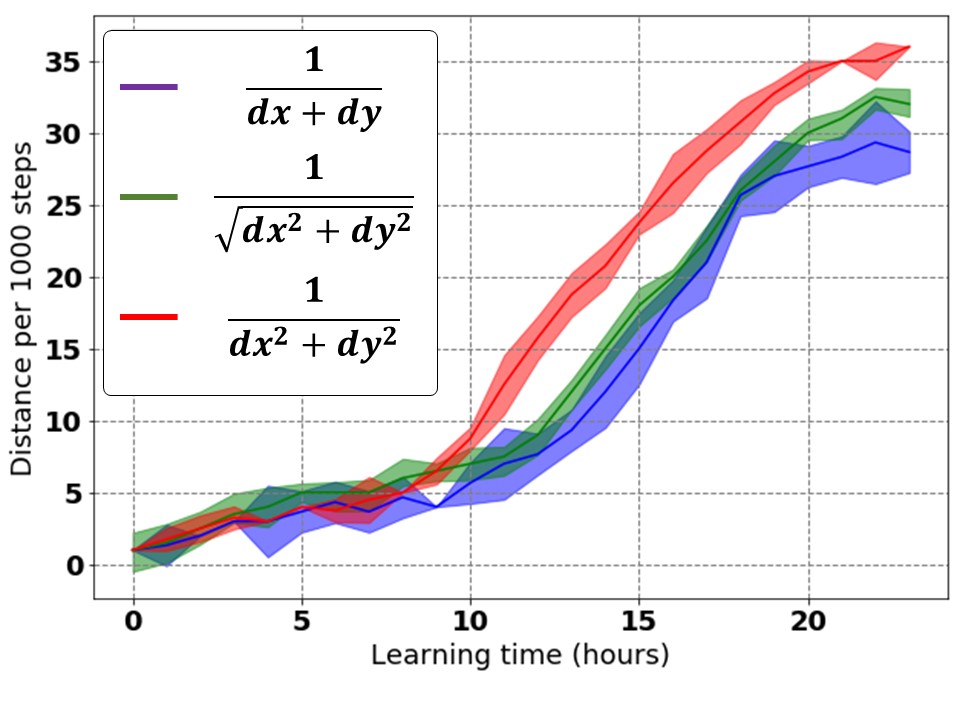}
\caption{Comparing three potential functions}
\label{img:var_potential}
\end{figure}

After selecting the running human video as the data source, we compared the three different potential functions as depicted in Figure \ref{img:var_potential}. The results show that PF3 performs best.

\section{Evaluation of Video-based Reward Shaping}

\begin{figure}
\includegraphics[width=0.47\textwidth]{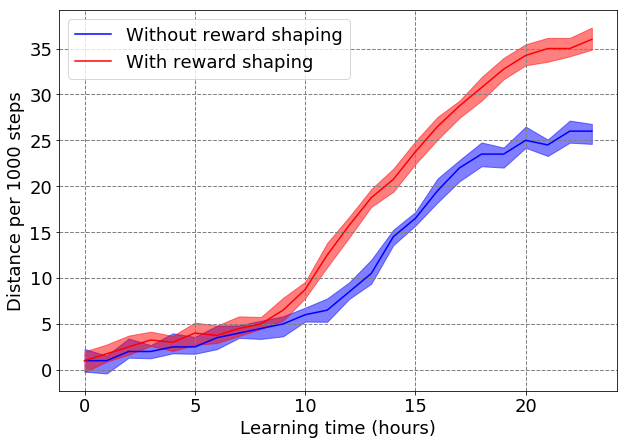}
\caption{Performance comparison between the baseline and the reward shaping approach}
\label{img:baseline_vs_shaping}
\end{figure}

Figure \ref{img:baseline_vs_shaping} shows the comparison of our chosen reward shaping approach (PF3) to the RL baseline. The results show that the reward shaping speeds up the learning significantly, reaching double the running speed at 12 hours of training. The end result after 24 hours of training still shows a significant advantage of the reward shaping approach. It is also worth noting that the demonstration video is of a running human who is using his arms, while the simulation model does not include these. 

\begin{figure}
\includegraphics[width=0.47\textwidth]{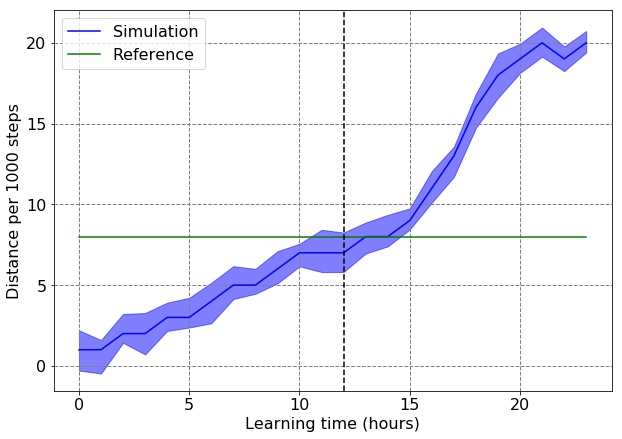}
\caption{Performance of the reward shaping approach with suboptimal data. The dotted vertical line represents 12 hours of training (the training time of the shaping source).}
\label{img:suboptimal}
\end{figure}

An important advantage of potential-based reward shaping is the theoretical guarantee that the shaping will not change the optimal policy. In order to demonstrate this advantage in our context,  we used a weak running robot generated by the baseline RL agent after 12 hours of training as a sub-optimal data source for the potential function. Clearly, the resulting agent is not running optimally, and the positions of the feet and knees will not be in optimal positions most of the time.  We then train our RL agent with the reward shaping generated from these sub-optimal coordinates (using PF3), and compared the performance to the weak runner. The results are shown in Figure \ref{img:suboptimal}, and demonstrate that the RL agent is able to overcome the suboptimal performance of the data source. In fact, after 20 hours of training, the performance is more than double that of the suboptimal running agent. Also, note that the suboptimal shaping did not hurt the learning performance significantly. After 12 hours of training the shaped agent performs comparable to the baseline agent with 12 hours of training.

\section{Conclusions}

In this paper, we presented a method to use videos of human and human-like running to shape the reward of an RL agent learning to run. Our results demonstrate that a significant improvement in learning speed can be achieved by our proposed method, as compared to a strong baseline which we designed combining selected techniques of the top ten entries to the "Learning to Run" competition at NIPS 2017. 

In future work, we intend to employ automated body pose extraction methods such as the one presented in \cite{guler2018densepose} and widen our investigation to other humanoid movement apart from running, e.g. jumping. 




\bibliographystyle{unsrt}
\bibliography{bibliography}

\end{document}